# Predicting Polymer Solubility in Solvents Using SMILES Strings

**Andrew Reinhard**

arbrf@umsystem.edu

**Thursday May 8th, 2025**

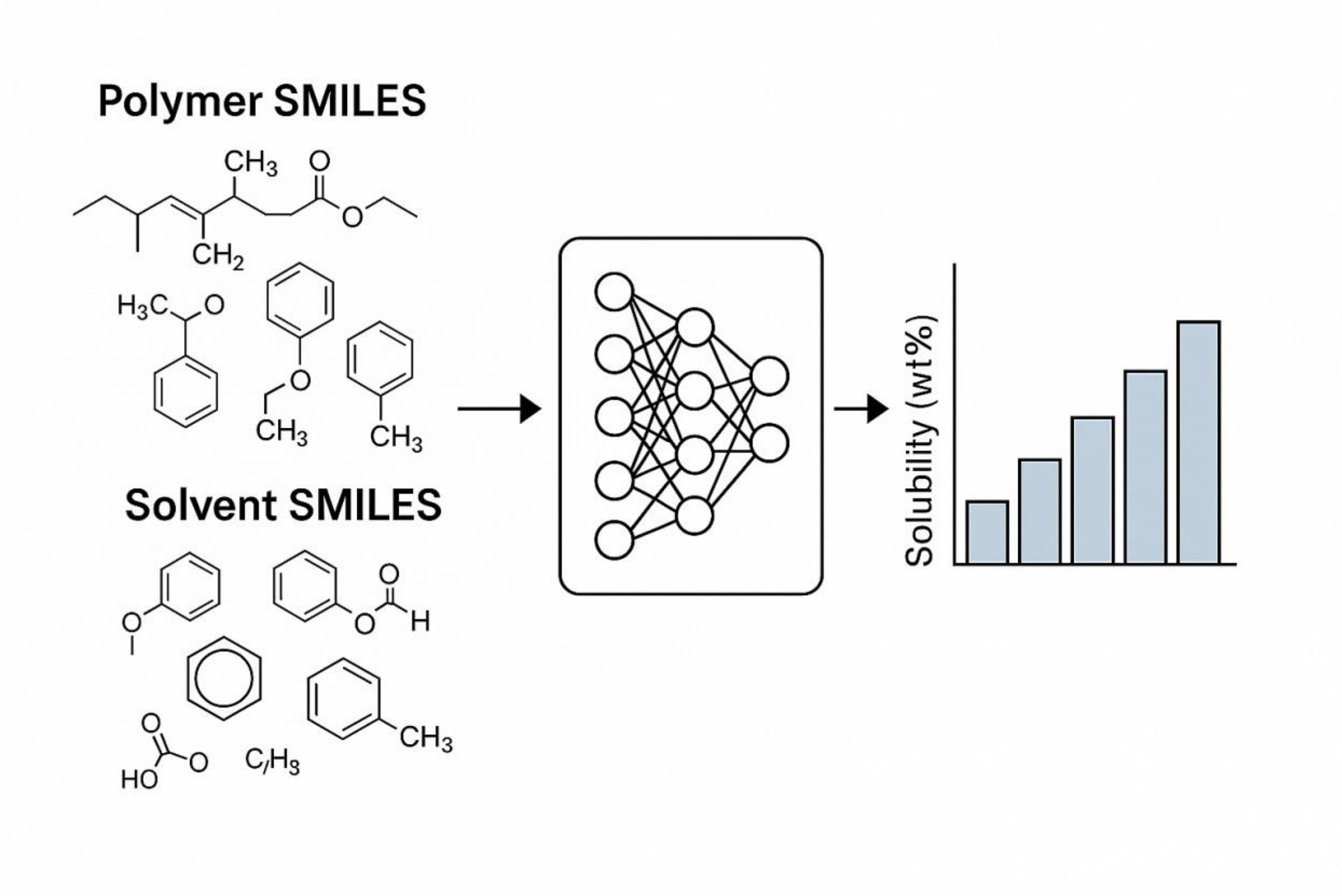

**Technical Description**

Understanding and predicting polymer solubility in various solvents is critical for applications ranging from polymer recycling to drug delivery. In this study, we develop a machine learning model that predicts polymer solubility, expressed in weight percent (wt%), directly from SMILES strings of both the polymer and solvent. The model is trained on a dataset of 8,049 polymer–solvent combinations at 25 °C, sourced from calibrated molecular dynamics simulations reported by Zhou et al. (2023). Eight commonly used polymers were included, and solvent SMILES strings were obtained using the PubChemPy library.[1] Molecular features were extracted through a combination of chemical descriptors (e.g., molecular weight, logP, TPSA) and fingerprints (1024-bit Morgan and 167-bit MACCS)[2,3], yielding a 2394-dimensional input vector per sample. A deep neural network with six hidden layers was trained using the Adam optimizer and evaluated using MSE loss. Model performance showed strong agreement between predicted and actual solubility values on both training and validation sets. Generalizability was assessed using experimental data from the Materials Genome Project, where the model maintained high accuracy on 25 unseen polymer–solvent pairs. These results demonstrate the viability of using SMILES-based deep learning models for accurate and scalable prediction of polymer solubility, offering potential for high-throughput solvent screening and green chemistry design.

## Introduction

The solubility of polymers in various solvents is a fundamental property that plays a critical role in a wide range of industrial and research applications, including plastics recycling, drug formulation, membrane fabrication, and polymer processing.[4–6] Accurate knowledge of solubility behavior enables informed solvent selection, improved material compatibility, and more sustainable design practices. However, experimental determination of polymer solubility is time-consuming, resource-intensive, and often infeasible for large-scale solvent screening.[7,8]

To address these limitations, computational approaches such as molecular dynamics simulations and group contribution methods have been developed to predict solubility behavior.[9,10] While these methods offer improvements in scalability, they often require extensive parameterization and domain-specific knowledge. Recent advances in cheminformatics and machine learning provide an opportunity to streamline solubility prediction by learning directly from molecular representations such as SMILES (Simplified Molecular Input Line Entry System) strings.[11,12] SMILES strings offer a compact and widely adopted format for representing molecular structures,[13] enabling rapid descriptor generation and high-throughput screening. When combined with machine learning, these representations can uncover complex relationships between chemical structure and solubility without the need for handcrafted rules or simulation frameworks.[14]

In this work, we present a deep learning model that predicts polymer solubility in organic solvents using only SMILES strings as inputs. The model is trained on data derived from validated MD simulations, focusing on eight industrially relevant polymers and over 1000 solvents at room temperature. We demonstrate strong performance on both the training set and external experimental validation data, highlighting the model's potential for use in solvent selection, polymer compatibility screening, and green chemistry applications.

## Methods

### *Data*

Data used to train this model were obtained from the Supporting Information of the paper by Zhou et al. (2023) in Green Chemistry.[9] The dataset includes solubility predictions for 8 common industrial polymers: EVOH (ethylene vinyl alcohol), PE (polyethylene), PP (polypropylene), PS (polystyrene), PET (polyethylene terephthalate), PVC (polyvinyl chloride), Nylon 6, and Nylon 66. For each polymer, solubility was predicted in 1007 different solvents at two temperatures: room temperature (defined as 25 °C) and an elevated temperature, which was set to one degree below the boiling point of each solvent.

For this study, only the room temperature data were used in order to improve relevance to standard laboratory conditions. The solubility data were obtained from molecular dynamics (MD) simulations calibrated using experimentally measured solubilities. These simulations were previously validated by the original authors and shown to be in good agreement with experimental data. Visualizations of the solutes and solvents are shown in Figure **1**, and were generated using RDKit.

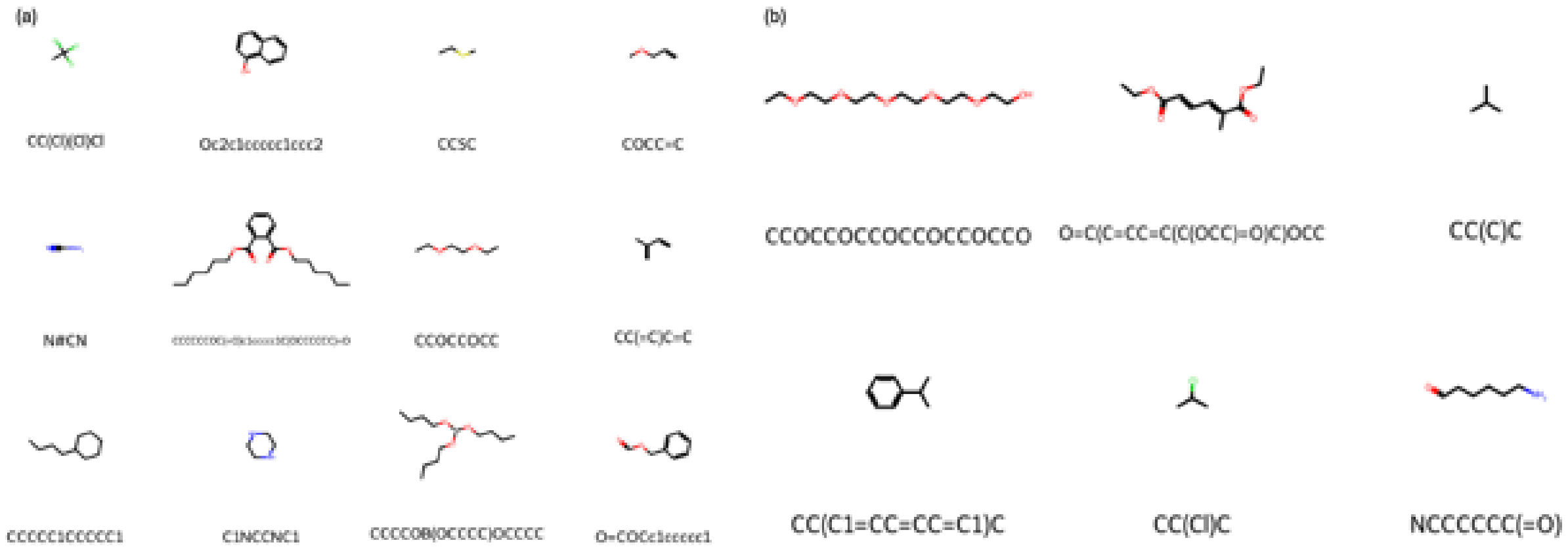

***Figure 1*** *Visualization of (a) solutes and (b) polymer solvents via RDKit*

The dataset includes the predicted solubility in weight percent (wt%) for each polymer–solvent pair, along with the CAS numbers for the solvents. Using the CAS numbers, SMILES strings for the solvents were retrieved via the pubchempy Python library. Polymer SMILES strings were represented using the shortest repeating unit for each polymer. The input features for the model were the SMILES strings of both the solvent and the polymer, along with the corresponding solubility (wt%). After filtering and cleaning the data, a total of 8049 unique polymer–solvent combinations were used for model training and evaluation.

*Data Preprocessing*

To generate features from SMILES strings, we extracted both molecular descriptors and chemical fingerprints for each solute and solvent. Specifically, we computed six molecular descriptors: molecular weight, logP, number of hydrogen bond acceptors, number of hydrogen bond donors, number of rotatable bonds, and topological polar surface area (TPSA). In addition, each molecule was represented using two types of fingerprints: a 1024-bit Morgan fingerprint and a 167-bit MACCS fingerprint. These descriptor and fingerprint vectors were concatenated to produce a 1197-dimensional feature vector per molecule. Combining the features from both the solute and solvent yielded a final input vector of 2394 features per sample.

*Model Architecture*

The model is a fully connected feedforward neural network that takes the 2394-dimensional input and outputs a single predicted solubility value (wt%). The architecture consists of six hidden layers with the following number of neurons: 2394 → 1024 → 512 → 256 → 128 → 64 → 32 → 1 (output). Each hidden layer is followed by a ReLU activation function, batch normalization, and a dropout layer with a dropout rate of 0.2 to prevent overfitting. The model was trained using the Adam optimizer with a learning rate of 1e-4 and a weight decay of 1e-5. The loss function used was Mean Squared Error (MSE). Training was performed with a batch size of 32, and early stopping was applied with a patience of 50 epochs to prevent overfitting. The target variable (wt%) was standardized using a StandardScaler before training and inverse-transformed after prediction to compute evaluation metrics.

*Baseline Comparison*

To evaluate the effectiveness of the neural network architecture, a baseline model using a Random Forest regressor was also trained on the same input features. Random Forests are widely used in cheminformatics due to their simplicity and robustness, and serve as a useful benchmark for assessing the value of deep learning.[15] While the Random Forest captured general trends in the data, it achieved significantly lower performance compared to the neural network. This comparison highlights the importance of using a deeper, more expressive model to capture the complex nonlinear relationships between polymer–solvent structure and solubility.

*External Validation*

To evaluate the generalizability of the model, it was tested using experimental data from the Materials Genome Project.[16] This database contains a wide range of polymers and solvents that have been experimentally characterized. One of the key measurements available is polymer mass fraction, which is equivalent to weight percent (wt%) divided by 100. The relevant data was scraped from the website. To ensure consistent validation conditions, only experiments conducted under atmospheric pressure were considered. Many of the tests involved repeated measurements for the same polymer-solvent combinations; these were averaged. Combinations with fewer than five tests were excluded from analysis. Ultimately, the model was validated on 25 unique solute–solvent pairs. A visualization of the validation data is shown in Figure 2.

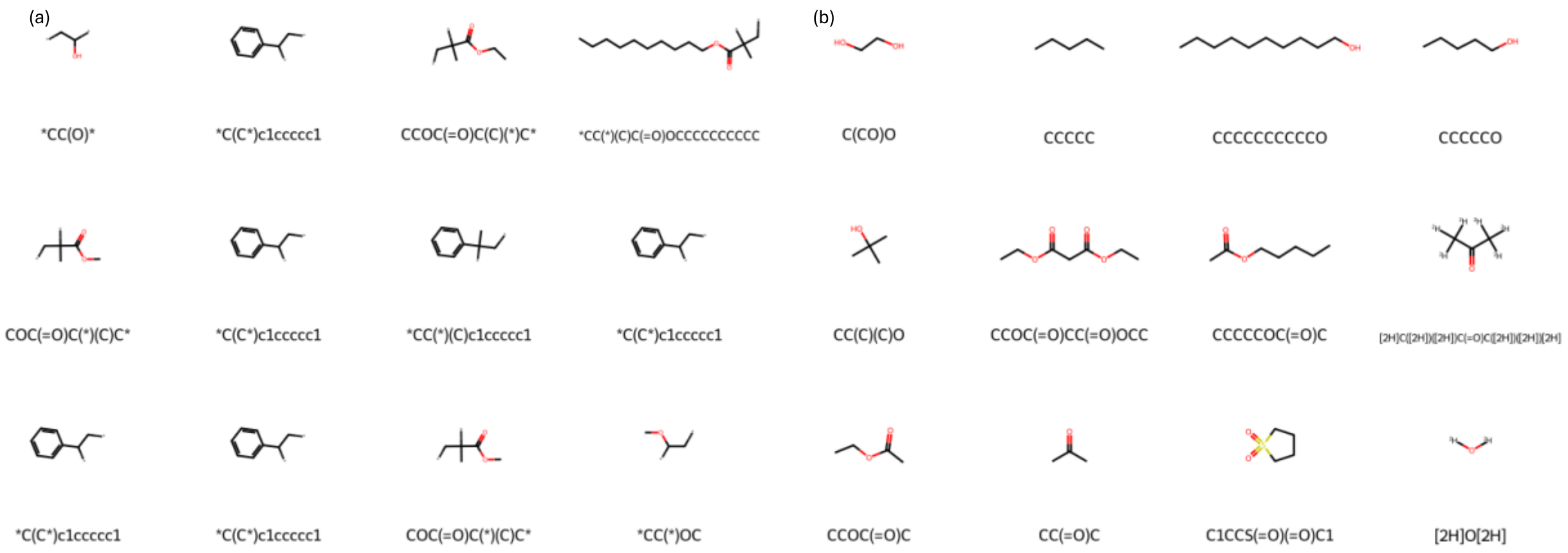

***Figure 2.*** *Visualization of the validation data (a) solutes and (b) polymer solvents via RDKit*

## **Experiments and Results**

*Baseline Model Performance*

To benchmark the deep neural network, a Random Forest regressor was trained using the same 2394-dimensional feature representation derived from SMILES-based descriptors and fingerprints. While Random Forest models are widely used in cheminformatics due to their robustness and low sensitivity to hyperparameter tuning, this implementation demonstrated clear performance limitations on the solubility prediction task.

Figure 3 shows the regression performance of the Random Forest model on the test set, where predicted solubility values (wt%) are plotted against the actual values. The regression line exhibits noticeable deviation from the ideal diagonal, with significant dispersion, particularly at higher solubility values. Quantitatively, the model achieved a mean absolute error (MAE) of 0.3946, a root mean squared error (RMSE) of 1.3620, and an $R^2$ of 0.4433. These metrics indicate that while the Random Forest model captured general solubility trends, it failed to learn the finer-grained nonlinear relationships necessary for high-accuracy prediction.

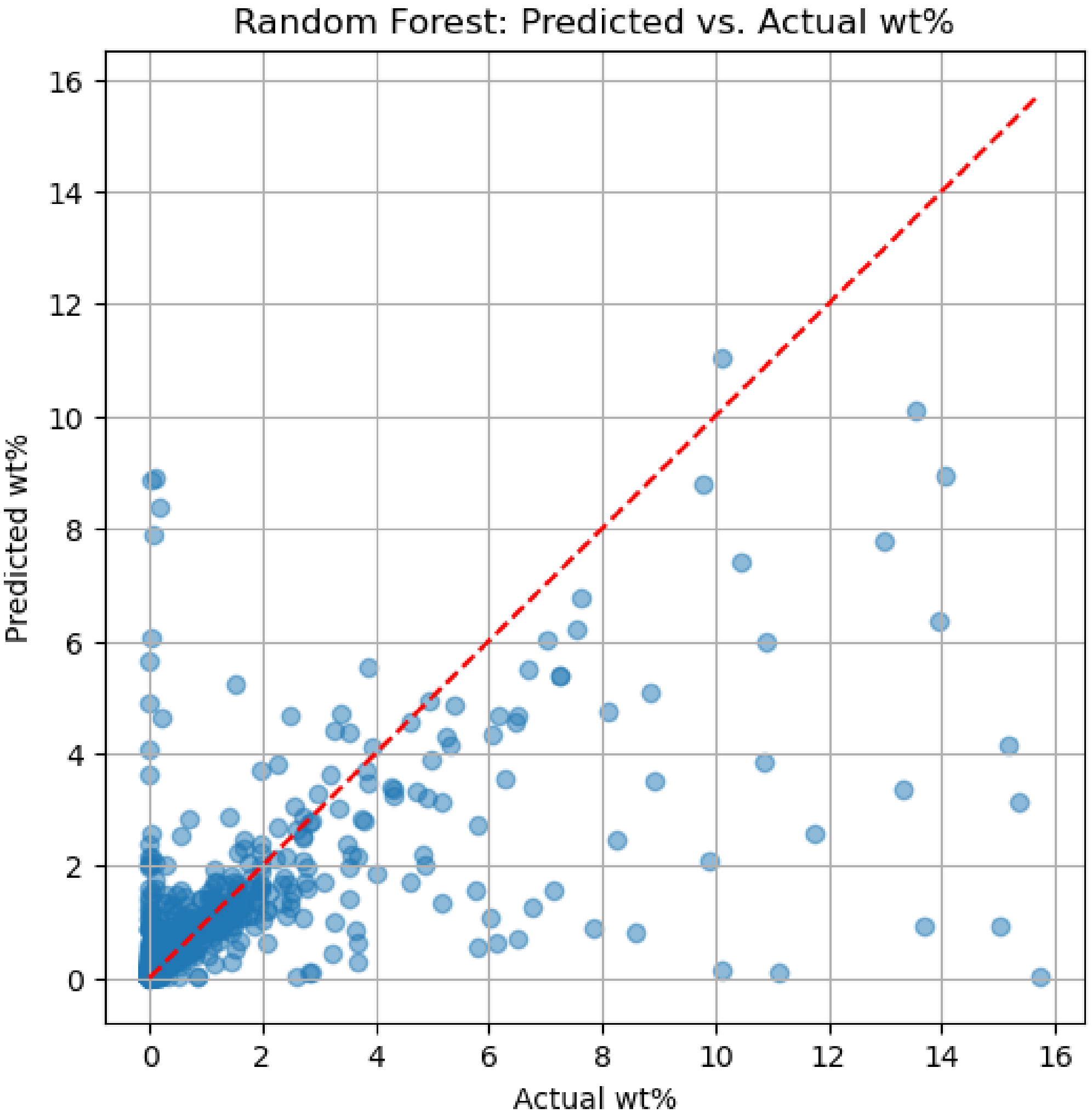

**Figure 3.** Visualization of the regression predicted vs actual for the validation data of the Random Forest

In contrast, the deep learning model achieved substantially lower error and over 80% explained variance ($R^2$ = 0.8076), more than doubling the performance of the Random Forest baseline. This reinforces the value of a deeper, more expressive architecture for mapping complex polymer–solvent interactions using only SMILES-derived features.

*Model Performance on Training and Validation Sets*

The deep learning model demonstrated strong predictive performance on both the training and validation datasets. Figure *4* shows the scatter plot of actual versus predicted solubility values (in wt%) on the validation set. The predicted values closely align with the diagonal line representing perfect prediction, indicating a high degree of correlation between the model outputs and the true values.

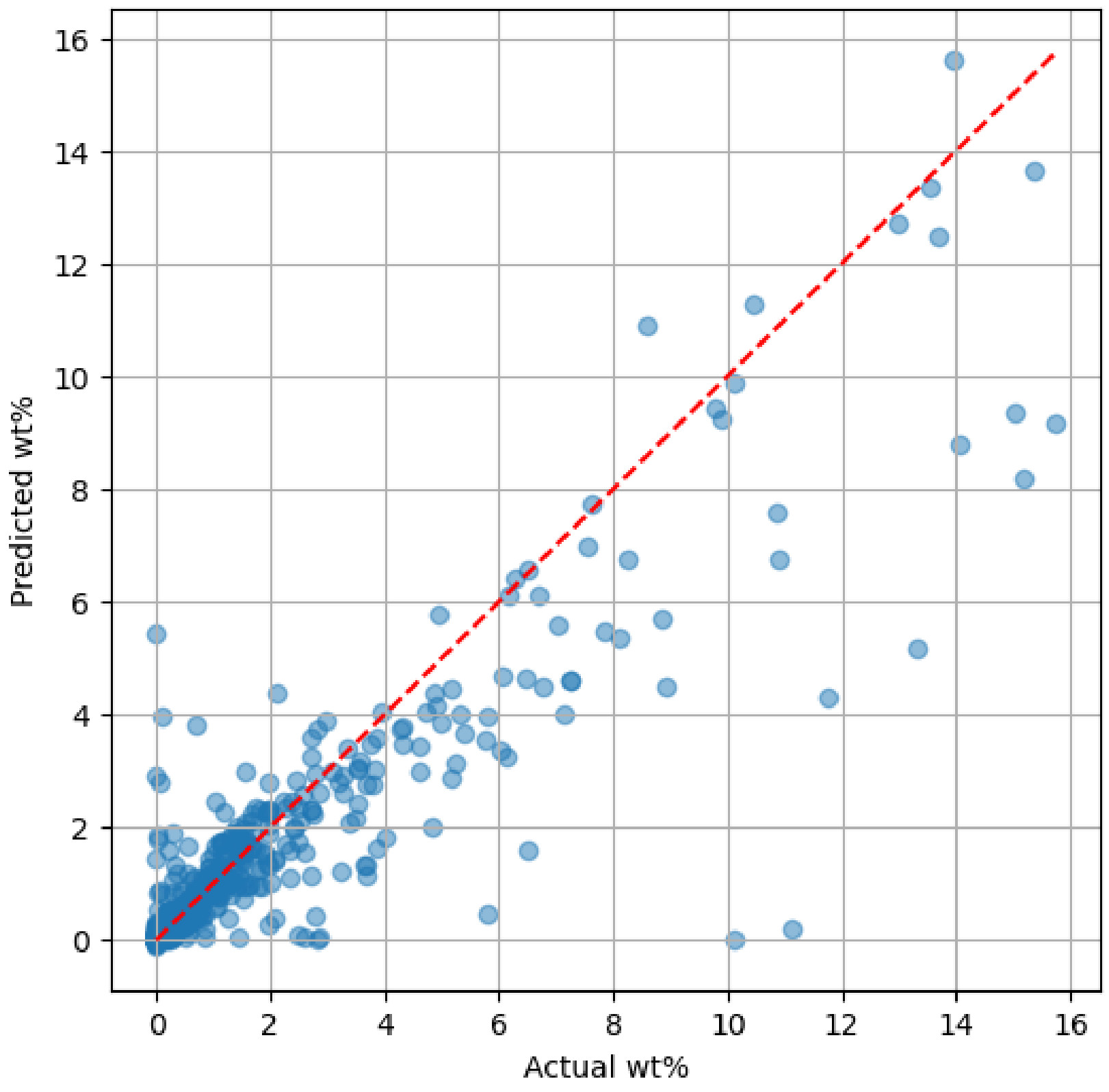

**Figure 4.** Visualization of the regression predicted vs actual for the validation data of the neural network

The model achieved a mean absolute error (MAE) of 0.2294, a root mean squared error (RMSE) of 0.8006, and a coefficient of determination ($R^2$) of 0.8076 across the combined training and validation sets. These metrics demonstrate the model's ability to learn complex, non-linear relationships between polymer–solvent molecular structure and solubility from the 2394-dimensional feature space derived from SMILES strings.

The relatively low MAE and RMSE suggest that the majority of solubility predictions fall close to their true values, while the $R^2$ value indicates that over 80% of the variance in the dataset is explained by the model. This level of accuracy is notable given the wide chemical diversity of the dataset, which spans over 1000 unique solvents and eight structurally diverse polymers. The results confirm that a deep neural network using only SMILES-derived features can serve as an effective and scalable tool for predicting polymer solubility in organic solvents, offering advantages in both speed and generalizability over traditional simulation-based methods.

*External Validation on Experimental Data*

To evaluate the generalizability of the model beyond the simulation-derived training data, we performed external validation using experimentally measured solubility data from the Materials Genome Project. This dataset includes polymer–solvent pairs that were not present in the training set, providing a rigorous test of the model's real-world applicability.

After filtering for consistency (e.g., atmospheric pressure, room temperature, and a minimum of five replicate measurements per pair), a total of 25 unique polymer–solvent combinations were selected for evaluation. For each pair, experimental solubility values were averaged across replicates to obtain a reliable ground truth for comparison.

Figure 4 shows the scatter plot of predicted versus actual solubility (wt%) for these 25 held-out combinations. The data points generally follow the expected diagonal trend, indicating strong agreement between the model's predictions and experimental observations. The model achieved an $R^2$ value of 0.7648 on this external dataset, reflecting a high level of predictive accuracy despite the domain shift from simulation to experiment. This highlights the generatability of the model to data that it was not trained on.

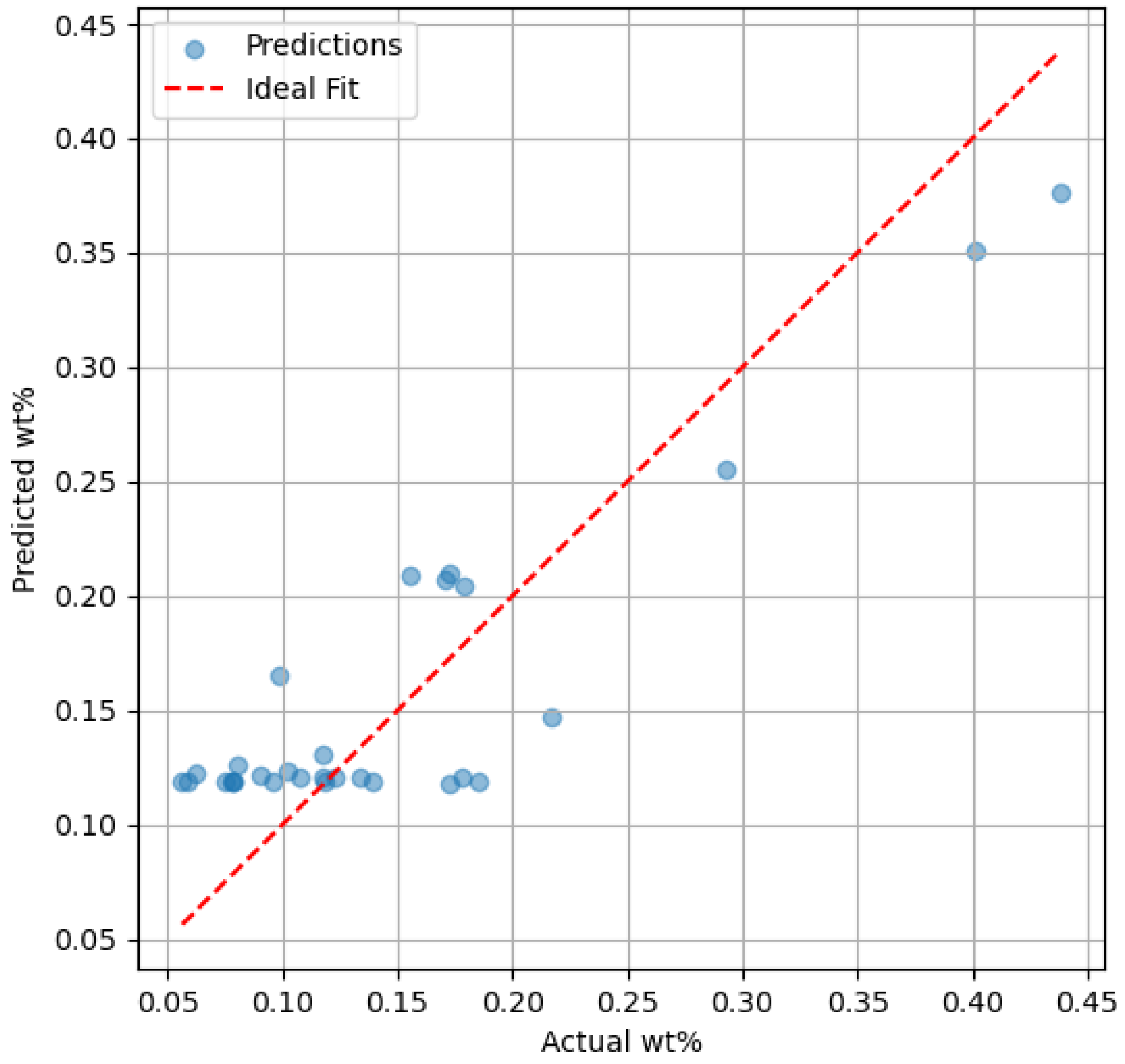

**Figure 5.** Saved models accuracy on experimental data outside of training set

While minor deviations are observed in a few cases, the overall trend confirms the model's robustness. Importantly, high accuracy was observed across different polymers, suggesting that the model maintains balanced performance across diverse polymer chemistries. This external validation underscores the potential of the SMILES-based neural network to serve as a practical screening tool for solvent selection and polymer processing applications, especially in contexts where experimental data are limited or unavailable.

*Solubility Classification Performance*

In addition to continuous solubility prediction, we evaluated the model's ability to perform binary classification of polymer–solvent pairs as either "good solvents" or "non-solvents." A threshold of 0.005 wt% was used to distinguish between the two classes, with solubility values greater than this cutoff labeled as good solvents, a cutoff used in polymer solubility literature to identify minimally effective solvent candidates.[15]

To perform this task, the model's output layer was adapted to include a sigmoid activation function, and binary cross-entropy was used as the loss function during training. Predictions were classified by thresholding output probabilities at 0.5. The classifier achieved an overall accuracy of 94.02% on the training and validation data, demonstrating excellent performance in distinguishing soluble from non-soluble combinations. The confusion matrix shown in Figure 6 highlights the low rates of both false positives and false negatives, with balanced classification across the two classes.

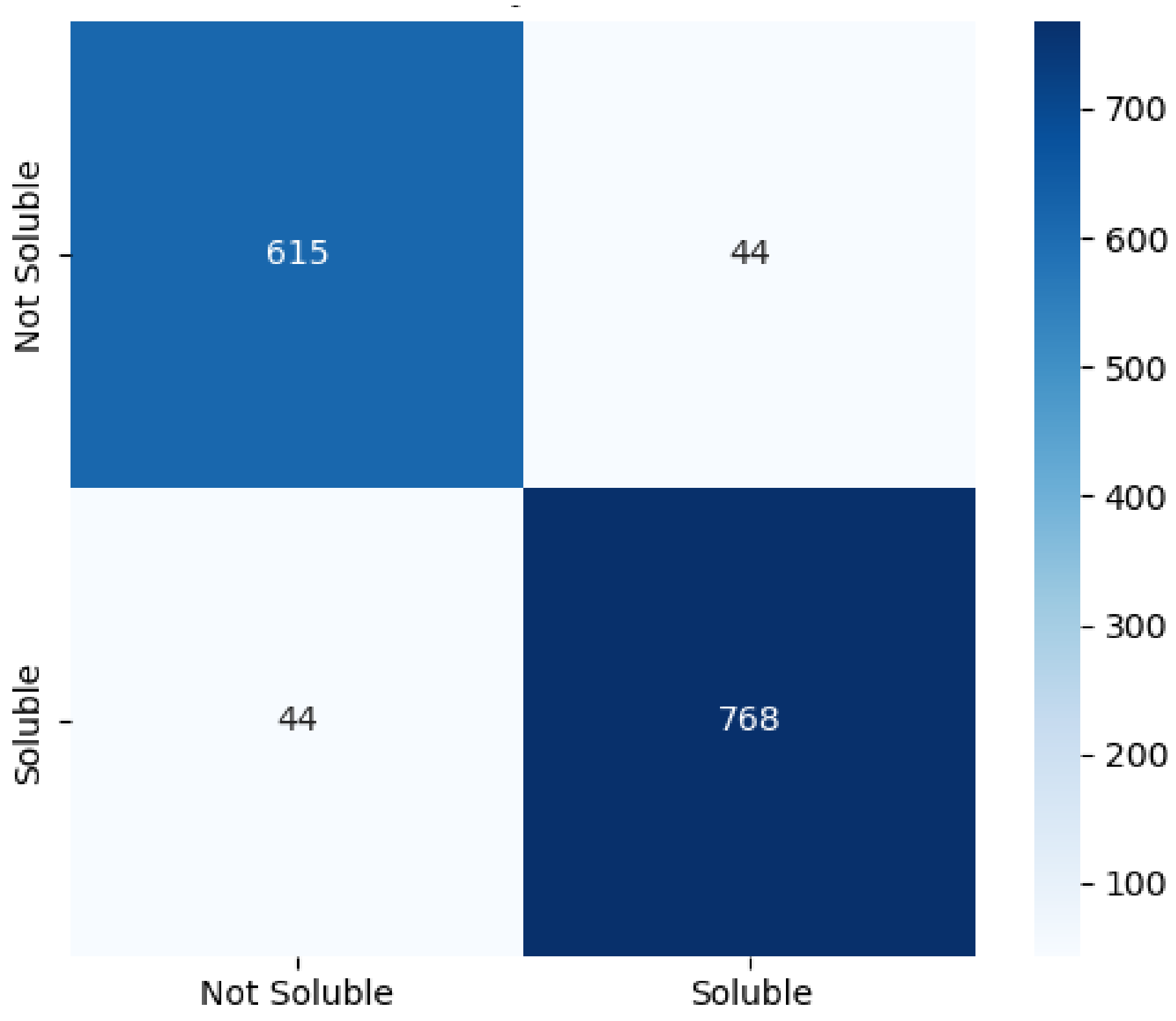

**Figure 6.** Classification Confusion matrix

It was not possible to evaluate this classification model on the external experimental dataset because all 25 polymer–solvent combinations in the experimental validation set were classified as soluble. As a result, the external data lacked examples of non-solvents, preventing direct validation of classification performance in a real-world setting. Nonetheless, the high accuracy observed on simulated data suggests strong potential for the model's application in high-throughput screening, where rapid elimination of poor solvent candidates is often desired.

## Conclusion

This study presents a deep learning framework for predicting polymer solubility in organic solvents using only SMILES strings as inputs. By combining molecular descriptors and chemical fingerprints for both polymers and solvents, we constructed a 2394-dimensional input representation that captures rich structural and physicochemical information. The model was trained on a dataset of 8049 polymer–solvent pairs derived from calibrated molecular dynamics simulations and demonstrated strong performance, achieving an $R^2$ of 0.8076, MAE of 0.2294, and RMSE of 0.8006 on the training and validation sets.

External validation using experimentally measured solubilities from the Materials Genome Project further confirmed the model's generalizability, with an $R^2$ of 0.7648 across 25 unseen polymer–solvent pairs. Additionally, the model achieved 94.02% accuracy in a binary classification task distinguishing good solvents from non-solvents, offering a fast and effective route for early-stage solvent screening. Although classification performance could not be externally validated due to the solubility of all experimental test cases, the results on simulation data demonstrate the model's potential for high-throughput compatibility assessments.

Overall, this work highlights the viability of SMILES-based deep learning approaches for solubility prediction, offering a scalable alternative to traditional simulation or experimental workflows. Future work may explore the inclusion of temperature-dependent behavior, model interpretability via explainable AI techniques, and expansion to broader polymer families and solvent conditions. These advancements could further support the development of data-driven tools for sustainable materials design and green chemistry innovation.

## Reflection

*Quick Summary*

This project aimed to develop a machine learning model capable of predicting the solubility of polymers in organic solvents using only SMILES strings as input. By combining molecular descriptors, MACCS keys, and Morgan fingerprints into a single high-dimensional representation, I trained a deep neural network to predict solubility in weight percent (wt%). The model was trained on simulation-derived data and validated on experimental results, achieving strong performance in both contexts. The model's ability to generalize beyond the training distribution, particularly to experimentally measured solubility values, suggests that this SMILES-based approach holds promise for practical solvent screening applications.

*Did everything turn out as expected?*

In many ways, the results exceeded my expectations. While I anticipated some alignment between predicted and experimental solubility values, the model's accuracy on external data was remarkably close to its internal validation performance. Achieving an $R^2$ of over 0.76 on the experimental dataset was particularly encouraging, especially considering the differences in data origin and measurement conditions. I had initially assumed that the model might struggle to generalize due to noise or structural differences in the experimental data, but it adapted well. This reinforced my confidence in the robustness of the combined fingerprint-descriptor representation.

There were also surprises. For example, the Random Forest baseline performed worse than expected, especially compared to the neural network. I had assumed tree-based models might offer competitive performance out-of-the-box, but this result clarified how essential deep architectures are for learning complex chemical relationships from high-dimensional inputs.

*Shortcomings*

Despite the successes, there are several limitations worth noting. First, the model was trained and tested only at room temperature (25 °C), which limits its ability to generalize to temperature-dependent solubility behavior. Future versions of the model should explicitly include temperature as an input variable to accommodate broader use cases. Second, the classification model—though highly accurate on internal data—could not be validated externally due to all the experimental pairs being soluble. This limited the ability to verify the classifier's ability to reject poor solvent candidates in real-world scenarios.

Another limitation lies in model interpretability. While correlation analysis provided some insight into feature relevance, the fingerprint-based inputs are inherently opaque. Without explicit tracking or feature attribution, it is difficult to explain exactly which molecular substructures or physicochemical patterns the model is relying on. As a result, the model functions largely as a black box, which could hinder its adoption in settings where explainability is critical.

Lastly, the dataset was drawn from simulations involving only eight polymers. While diverse in structure, this is still a relatively narrow slice of polymer chemistry. There's no guarantee the model will perform well on novel chemistries outside this training distribution, especially biopolymers or complex copolymers.

*Future Work*

Several promising avenues exist for extending this work. First, I would expand the dataset to include more diverse polymer families, particularly those relevant to sustainability, biomedical engineering, or energy applications. Increasing the chemical diversity would not only improve generalizability but also allow for more nuanced classification tasks, such as predicting partial solubility or phase behavior.

Second, incorporating temperature and pressure as explicit input variables would make the model more useful for real-world processing and manufacturing workflows. A multi-task learning approach might also be explored to simultaneously predict solubility and other polymer–solvent properties like diffusion or swelling.

Finally, this model could be embedded in a larger pipeline for green chemistry or circular materials design. For example, it could be used to identify environmentally benign solvent replacements or to screen for optimal dissolution strategies in plastic recycling workflows. Paired with optimization algorithms, the model could support inverse design applications, recommending solvent–polymer systems based on target solubility constraints.